\documentclass[conference]{IEEEtran}
\usepackage{url}
\usepackage{booktabs}
\usepackage{tikz}
\usepackage{amsmath}
\usepackage{graphicx}
\usepackage{float}
\usepackage{fancyhdr}
\usepackage{xcolor}

\usetikzlibrary{arrows.meta, positioning, shapes.geometric}
\fancypagestyle{firstpage}{
  \fancyhf{} 
  \fancyfoot[C]{Your Footer Here} 
}
\begin{document}
\title{Adversarial Machine Learning Threats to Spacecraft}
\thispagestyle{firstpage}

\makeatletter
\newcommand{\linebreakand}{%
  \end{@IEEEauthorhalign}
  \hfill\mbox{}\par
  \mbox{}\hfill\begin{@IEEEauthorhalign}
}
\makeatother

\author{
  \IEEEauthorblockN{1\textsuperscript{st} Rajiv Thummala}
  \IEEEauthorblockA{\textit{Sibley School of Mech. and Aerospace Engineering} \\
    \textit{Cornell University}\\
    Ithaca, NY USA \\
   rkt34@cornell.edu}
  \and
  \IEEEauthorblockN{2\textsuperscript{nd} Shristi Sharma}
  \IEEEauthorblockA{\textit{Department of Computer Science} \\
    \textit{The University of North Carolina at Chapel Hill}\\
    Chapel Hill, NC \\
    ssharma@unc.edu}
  \linebreakand 
  \and
  \IEEEauthorblockN{3\textsuperscript{rd} Matteo Calabrese}
  \IEEEauthorblockA{\textit{Department of Systems Engineering} \\
    \textit{Cornell University}\\
   Ithaca, NY USA \\
    mc2884@cornell.edu}
  \and
  \and
  \IEEEauthorblockN{4\textsuperscript{th} Gregory Falco}
 \IEEEauthorblockA{\textit{Sibley School of Mech. and Aerospace Engineering} \\
    \textit{Cornell University}\\
    Ithaca, NY USA \\
    gfalco@cornell.edu}
}


\IEEEoverridecommandlockouts
\makeatletter\def\@IEEEpubidpullup{6.5\baselineskip}\makeatother

\maketitle

\begin{abstract}
Spacecraft are among the earliest autonomous systems. Their ability to function without a human in the loop have afforded some of humanity's grandest achievements. As reliance on autonomy grows, space vehicles will become increasingly vulnerable to attacks designed to disrupt autonomous processes - especially probabilistic ones based on machine learning. This paper aims to elucidate and demonstrate the threats that adversarial machine learning (AML) capabilities pose to spacecraft. First, an AML threat taxonomy for spacecraft is introduced. Next, we demonstrate the execution of AML attacks against spacecraft through experimental simulations using NASA’s Core Flight System (cFS) and NASA's On-board Artificial Intelligence Research (OnAIR) Platform. Our findings highlight the imperative for incorporating AML-focused security measures in spacecraft that engage autonomy.
\end{abstract}

\section{Introduction}
The space domain has undergone a significant paradigm shift, evolving from a predominantly exploratory and scientific domain into a strategic and contested sphere, integral to national security and defense postures globally. This transformation is underscored by the recognition of space as a critical theater for military operations, where the assets deployed—ranging from communication and navigation satellites to surveillance and reconnaissance platforms—constitute pivotal elements in the infrastructure of modern warfare. As nations increasingly depend on these assets for both strategic advantage and operational capabilities, the imperative to safeguard them from a spectrum of threats has never been more paramount.

Concurrently, the integration of artificial intelligence (AI) within space systems has been accelerating, driven by the need for enhanced data processing, autonomous decision-making capabilities, and improved operational efficiency. This reliance on AI systems marks a significant evolution in spacecraft design and functionality, enabling more sophisticated missions and responsive space operations. However, the incorporation of AI also introduces new vulnerabilities, particularly in the context of adversarial machine learning (AML). AML techniques, which are designed to manipulate or deceive AI models through carefully crafted inputs, represent a burgeoning threat vector that spacecraft systems are ill-prepared to counteract.

The hardening of spacecraft against cyber threats, while nascent, has been an area of growing focus within the aerospace defense sector. Yet, the threats posed by AML specific to spacecraft have not been adequately explored or addressed, leaving a critical gap in risk, threat, and vulnerability assessments for spacecraft engineers. The sophistication and adaptability of AML attacks exacerbate this oversight, as traditional cybersecurity measures prove insufficient against threats that are designed to exploit the inherent vulnerabilities of AI systems. This discrepancy underscores the urgent need for a concerted effort to understand the unique implications of AML in the space domain, to inform spacecraft engineers and designers in effectively diagnosing and mitigating these emerging threats.

Accordingly, this paper details a taxonomy of AML threats to spacecraft, with emphasis on the dynamics between mission-specific objectives, the operational environment, constraints of spacecraft, and the inherent vulnerabilities introduced by the integration of machine learning capabilities. By systematically categorizing these threats based on identified parameters, we offer a structured approach to facilitate the assessment and comprehension of potential AML attacks that spacecraft may encounter throughout their lifecycle. Furthermore, to demonstrate the feasibility of such attacks and inform our insight, we execute a series of AML attacks against a simulation of a spacecraft utilizing NASA's Core Flight System and the NASA On-board Artificial Intelligence Research (OnAIR) Platform. Finally, we outline the implications of AML attacks on spacecraft, mitigations, and next-steps. 

\section{Background and Prior Art}
\subsection{Adversarial Machine Learning Overview}
Machine learning (ML), a field within artificial intelligence, characterizes the ability of computers to learn from provided data without being explicitly programmed for a particular task. Adversarial machine learning (AML) is the process of extracting information about the behavior and characteristics of a ML system and/or learning how to manipulate the inputs into a ML system in order to obtain a preferred outcome \cite{nistArtificialIntelligence}. This process is deliberately employed by adversaries to deceive ML models, circumvent or impair systems, mislead automated decision-making processes, and corrupt data integrity in applications.

In the context of AML, the manipulated inputs designed to subvert ML algorithms are known as "adversarial examples". Adversarial examples represent a formidable threat to both information technology (IT) and operational technology (OT) systems. Hardening OT systems (embedded systems) against AML attacks is especially imperative, given their cyber-physical nature and consequent designation as safety-critical systems. In 2021, researchers demonstrated that they could cause an autonomous vehicle's computer vision system to mistake a stop sign for a speed limit sign by putting a small, almost imperceptible sticker on the sign \cite{nistNISTArtificial}. A year later, researchers at MIT successfully deceived an image classifier, causing it to misclassify machine guns as helicopters. They argued that if a weapons system were equipped with computer vision and trained to neutralize specific machine guns, such misidentification could lead to unintended passivity, potentially creating life-threatening vulnerabilities in the computer's ML algorithm \cite{afceaAdversarialMachine}. Apart from crafting adversarial examples, AML also encompasses traditional integrity violations to ML models. Akin to tampering with source code of software to enable malicious operations, adversaries may modify pre-trained models to degrade its functionality or leak data. 

\subsection{Classifications of Adversarial Machine Learning}
Attacks against machine learning models are typically categorized into three distinct categories: black-box, white-box, and transfer attacks. These categories delineate the extent of an attacker's knowledge about the target model, ranging from limited to complete model familiarity. Each category significantly impacts the kill chain an attacker undertakes to compromise the target system. 

\subsubsection{White Box Attacks}
White Box attacks in the context of AML occur when the attacker has complete knowledge of the machine learning model, including its architecture, parameters, and training data. This comprehensive understanding enables attackers to craft adversarial examples with high precision, making these attacks particularly potent. Common white box attacks include poisoning attacks, model inversion attacks, and backdoor attacks. 
\begin{enumerate}
    \item Poisoning Attacks: Attacker corrupts the training process by introducing malicious data into the training set. This can lead to the model learning incorrect patterns, which affects its performance during the inference stage.
    \item Model Inversion Attacks: Attacker aims to reverse-engineer the model's inputs or training data by extracting sensitive information about the dataset or the features used by the model, potentially compromising data privacy.
    \item Backdoor Attacks: Attacker inserts a backdoor into a model during training. When the model encounters a specific trigger input during inference, it produces an incorrect output, while functioning nominally for other outputs.
\end{enumerate}

\subsubsection{Black Box Attacks}
In contrast to White Box attacks, Black Box attacks are executed without knowledge of the model’s internal workings. Attackers typically have access only to the model’s inputs and outputs. These attacks are more representative of real-world scenarios where internal details of a system are not readily accessible. Common black box attacks include evasion attacks, model extraction attacks, and membership inference attacks. 
\begin{enumerate}
    \item Evasion Attacks: Attackers craft misleading input based on the model's output without knowing the specific details of the model
    \item Model extraction Attacks: Attacker queries the model and observes its outputs to approximate its functionality without direct knowledge of its internals. 
    \item Membership Inference Attacks: Attacker observes the behavior of a target ML model and predicts examples that were used to train it. After gathering enough high confidence records, the attacker uses the dataset to train a set of "shadow models" to predict whether a data record was part of the target model's training data. This information is exploited to execute succeeding attacks.  
\end{enumerate}

\subsubsection{Transfer Attacks}
Transfer attacks involve developing attacks on a surrogate model (which the attacker has access to) and then transferring these attacks to the target model. Adversarial examples generated for one model can often deceive another model, even if the two models have different architectures or were trained on different datasets.  Common transfer attacks include adversarial reprogramming and evasion attacks:
\begin{enumerate}
    \item Adversarial Reprogramming: Attacker leverages knowledge from one machine learning model to manipulate the inputs of another model, re-purposing it for a task it wasn't originally designed to perform, without changing its internal architecture or training.
    \item Evasion Attacks: Attacker creates adversarial examples using one model and applies them to deceive another, exploiting similarities in how both models process inputs to induce incorrect predictions or classifications.
\end{enumerate}

Transfer evasion attacks fool the model into making errors under specific conditions, while adversarial reprogramming is consistently using the model for a completely different task. Thus, evasion deals with deception on a case-by-case basis, whereas adversarial reprogramming repurposes the model's capabilities.

\subsection{Existing Studies}
There is a considerable lack of literature pertaining to the impact of AML on spacecraft. The Aerospace Corporation's Space Attack Research and Tactic Analysis (SPARTA) matrix\cite{spartaMATRIX}, the leading framework for spacecraft threat assessments, touches at a high level on poisoning AI/ML training data as a potential tactic, technique, and procedure (TTP), but does not consider it appropriately in the context of spacecraft. \cite{aerospacePoisonAIML}.

Rather than focusing on the primary aspect of AML concerning spacecraft—misleading on-board computer vision systems into misclassifying data through simple alterations of the target's physical features—SPARTA limits its consideration of AML to a means for reducing the effectiveness of threat actor attack detection, presumably in the context of edge-based security mechanisms such as intrusion prevention systems (IPS) and intrusion detection systems (IDS). Additionally, SPARTA does not explicitly differentiate between the risks posed by traditional ground-based training operations conducted prior to loading the model onto the spacecraft and the unique vulnerabilities introduced by edge-based (online) training, which is becoming a standard approach. The framework also fails to highlight what spacecraft design choices and parameters are likely to heighten vulnerability to AML, instead enumerating generic AML countermeasures such as implementing ML data integrity, process white listing, data backup, IPS/IDS implementation, etc. These limitations prevent SPARTA from adequately informing spacecraft engineers about the spectrum of AML threats their spacecraft are likely to encounter, leading to an increased vulnerability to AML-related security risks. Other recently published standards such as NIST's Introduction to Cybersecurity for Commercial Satellite Operations \cite{nistIR8270} in July 2023 fail to even acknowledge AML as a potential threat to spacecraft.

Despite the abundance of literature on AML at large, research specifically addressing its impact on spacecraft remains scant. One of the few studies that attempts to bridge this gap focuses on resilient machine learning in space systems, demonstrating adversarial attacks and defenses against a pose estimation algorithm \cite{ieeeResilientMachine}. However, it noticeably lacks depth with regards to spacecraft systems engineering. Notably absent is an analysis of attack vectors relevant to spacecraft, as well as an analysis of how specific design decisions for spacecraft may render them susceptible to AML threats. In addition, while the paper applies AML techniques against a space-related dataset, the experiments are not conducted within a simulated spacecraft flight software environment, limiting the practical applicability of the findings to real-world spacecraft missions. Ultimately, the research offers limited utility for spacecraft engineers, as it provides no guidance on conducting threat assessments and focuses solely on software-based mitigations (adversarial defenses) without considering the comprehensive attack surface of a spacecraft,  such as ground segment intrusions and supply chain compromises. Other papers such as \cite{Du_Chen_Chin_Law_Sasdelli_Rajasegaran_Campbell_2022} explore the subject of adversarial patches against aerial imagery, which is of use for attacking on-board image classifiers. However, the process of injecting the attack into the spacecraft's software bus or on-board computer is not discussed, which is a major missing gap in available literature. These critiques are equally applicable to \cite{Du_Law_Sasdelli_Chen_Clarke_Brown_Chin_2022} and \cite{Hickling_Aouf_Spencer_2023} which present robust attacks, but are not demonstrated in a spacecraft's flight software environment, hindering external validity for spacecraft engineers. 

There accordingly exists a critical gap in research for developing a comprehensive threat taxonomy pertaining to AML risks to spacecraft. The following content provides a non-exhaustive analysis of AML threats to spacecraft with the intention to remediate voids in existing literature and frameworks.

\section{AML Threat Taxonomy for Spacecraft}
The threat of AML attacks faced by a spacecraft is present at all stages of its life cycle, panning  system design, assembly, integration, testing, launch, operational phases, and decommissioning. Two broad classes of AI systems and functionalities are employed by spacecraft, predictive and generative AI, both of which contrast in regards to their threat matrix. Predictive AI employs statistical algorithms and machine learning to analyze existing data and make predictions or forecasts based on patterns and trends \cite{nistNISTArtificial}. Generative AI, on the contrary, is employed for creating new content or data based on input data. 

Predictive AI encompasses the overwhelming majority of ML functions performed by spacecraft due to its aptitude for pattern recognition and decision making. For instance, edge-based image classification algorithms (computer vision) and autonomous navigation systems are enabled utilizing predictive AI techniques. Spacecraft engaged in space traffic management, however, may employ generative AI (e.g. generative adversarial networks (GANs)) to perform simulations and risk assessments of potential maneuvers to avoid conjunctions. Both predictive and generative AI systems are vulnerable to adversarial examples, albeit through distinct mechanisms. Predictive AI systems are predominantly exposed to evasion attacks, designed to mislead models into rendering inaccurate predictions or classifications \cite{nistNISTArtificial}. Conversely, generative AI systems frequently encounter data poisoning attacks, wherein the training data is deliberately manipulated to degrade the quality of the model's generated outputs \cite{nistNISTArtificial}. The taxonomy of AML threats faced by a spacecraft is predicated on the application of predictive AI versus generative AI in addition to the following parameters: 

\begin{enumerate}
    \item Mission Objectives and Operational Context
    \item Resource Constraints
    \item Learning Architecture, Method, and Stage
    \item Storage Architecture
    \item Command and Data Handling (C\&DH) Accessibility 
    \item Model Exposure and Interaction
\end{enumerate}
\subsection{Parameters}

\subsubsection{Mission Objectives and Operational Context}
The specific mission objectives and operational context of a spacecraft are key determinants in shaping its application of AI. These factors not only guide the class of AI (predictive vs. generative) and its deployment strategy, but also significantly influence the spectrum of AML attacks that the spacecraft is likely to face. For example, in the case of a Falcon 9 rocket autonomously landing on an autonomous spaceport drone ship (ASDS), the primary AML threat would be to the spacecraft's computer vision system used for guidance. Attackers might attempt evasion attacks by intelligently altering the drone ship's visual markers (e.g., changing patterns, introducing noise, etc.), misleading the computer vision algorithms and potentially leading to landing errors. In contrast, model inversion attacks aimed at extracting sensitive data from the model’s training set, which primarily pose a risk to data confidentiality, hold minimal relevance for the operational aspects of the Falcon 9's landing procedure. In such instances, the overarching concern is the preservation of operational integrity rather than the protection of training data confidentiality.

\subsubsection{Resource Constraints}
Spacecraft are inherently resource constrained due to size, weight, and power (SWaP) restrictions coupled with design constraints from remote operation in a harsh and inaccessible environment. This characteristic significantly influences the types of AML attacks an adversary might conduct against a spacecraft, due to the resource-intensive demands of machine learning. 

Adversaries may choose attacks that exploit the limited processing power and overhead capacity of spacecraft. For instance, complex evasion attacks requiring the AI system to process subtly altered but computationally intensive inputs could strain the system, leading to slowed response or system overloads. The limited storage capacity of a spacecraft necessitates that edge-based AI systems must employ simplified models or reduced datasets, which could be less robust against certain types of attacks. An adversary may exploit this property by designing attacks that a more robust model could have countered but a lightweight model cannot. For instance, if the system is optimized for speed over accuracy due to processing limitations, an adversary may use evasion techniques that exploit this trade-off. 

An example is a spacecraft equipped with an edge-based AI system responsible for autonomous navigation and obstacle avoidance in an environment dense with debris or in close proximity to other spacecraft. The AI system would be designed to prioritize speed over accuracy in its computations to ensure that the spacecraft can make rapid adjustments to its trajectory. This is crucial in space, where the relative velocities of objects can be extremely high, and the window for evasive maneuvers is limited. An adversary, aware of the AI system's preference for speed over accuracy, may deploy a series of complex evasion attacks. This could involve subtly altering the characteristics of the objects in the spacecraft's path (e.g., through spoofing or jamming signals) to make them appear differently to the spacecraft's sensors. These alterations would be designed to be computationally intensive to analyze accurately, exploiting the system's simplified models or reduced datasets that are not robust against such nuanced attacks.

Edge-based intrusion prevention systems on spacecraft are constrained by overhead limitations, which impede their ability to detect and respond accurately to model tampering. Additionally, the limited power supply on spacecraft makes AML attacks that drain energy more viable. By inducing the edge-based AI system to engage in repeated and unnecessary computations an attacker could ultimately drain the spacecraft's power supply. For instance, adversaries with C\&DH access would be able to increase the number of epochs to levels beyond the spacecraft's processing capabilities, effectively leading to a denial of service. Given the significant overhead required to update or retrain models on the edge of spacecraft, adversaries are likely to opt for attacks that exploit known vulnerabilities which cannot be swiftly patched or updated, making persistent or slowly evolving attacks more attractive. If a spacecraft has limited down-linking capacity for analysis, adversaries may employ attacks that corrupt or alter data in a way that is not immediately detectable, exploiting the delayed or limited verification process.

\subsubsection{Learning Architecture, Method, and Stage}
The learning architecture, method, and stage significantly impact the primary attack vectors adversaries will exploit to perform AML against a spacecraft. Spacecraft that pre-load trained models, for example, are especially susceptible to white box attacks due to increased accessibility to the training process. Adversaries may penetrate sensitive IT systems and communication channels in the ground segment to gain insights into the spacecraft's model and supply chain, identifying potential vulnerabilities for exploitation. Reliance on pre-loading models circumvents the need for adversaries to infiltrate the ground station for uplink access, markedly simplifying efforts to compromise the model's integrity. 

Traditionally, spacecraft employing AI avoid training models on the edge of the vehicle due to the significant computational overhead cost, opting to pre-load models prior to launch. This paradigm has begun to shift, however, with vehicles training directly on the edge due to advancements in edge computing and novel learning architectures for space systems. For instance, \cite{10121575} discusses the application of federated learning for multi-agent space systems (i.e.: satellite constellations) to decentralize the training process across multiple spacecraft instead of one vehicle bearing the entire computational overhead. Edge-based online learning capabilities for spacecraft are expected to standardize in the coming decades and will transform the AML threat landscape for spacecraft. 

Spacecraft training on the edge, while equally vulnerable to adversarial examples, would likely be targeted by bespoke AML attacks that exploit characteristics inherent in spacecraft. As aformentioned, computing constraints could be exploited by deploying malware that drastically increases the number of epochs of a model to cause computational exhaustion. Those that employ federated learning as their training architecture would likely be targeted with model poisoning or backdoors, whereby a participating agent could crosslink an infected portion of the model. When adversaries possess C\&DH access, they are unlikely to prioritize subtler AML attacks such as evasion, inversion, or extraction. This is because direct manipulation through write access offers a more powerful and immediate means of compromise, rendering traditional querying-based attacks less attractive in comparison. 

The learning stage (training-time, inference time, etc.) and method (unsupervised learning, supervised learning, reinforcement learning, etc.) are general indicators for what type of AML attack an adversary may select regardless of the type of target (IT systems, OT systems, etc.). NIST AI 100-2 E2023 \cite{nistNISTArtificial} highlights various attacks that are optimal for deployment with regards to these variables. In addition to referencing this publication, spacecraft engineers must take into account the nuances of spacecraft flight software (FSW) development. For instance, FSW developers extensively deal with COTS components that are effectively black boxes. As stated by CISA, “they [COTS customers] can review neither the code nor the architecture. In general, COTS buyers have to rely on the reputation of the developers, published security reports, and security forums” \cite{CISA}. FSW developers may not have access to the source code of a model employed by a COTS product and will therefore be at a disadvantage in remediating vulnerabilities. In fact, unless further insights can be gained through documentation, they may not be informed about the learning processes or methods employed, significantly exacerbating the ability to implement mitigations.

\subsubsection{Storage Architecture}
Spacecraft employ various storage architectures dependent on their mission requirements. These storage architectures play a crucial role in supporting the operation of machine learning models on board, which accordingly influences the range of AML threats to which a spacecraft may be vulnerable. Two primary storage methods are utilized by spacecraft: Electronically Erasable Programmable Read-Only Memory (EEPROM) and solid-state drives (SSDs). The distinctions between EEPROMs and SSDs from a ML perspective are highlighted in table 1. 

\begin{table}[htbp]
\centering
\caption{Comparison of EEPROM and SSD for ML in Spacecraft}
\begin{tabular}{p{0.15\columnwidth}p{0.35\columnwidth}p{0.35\columnwidth}}
\toprule
\textbf{Aspect} & \textbf{EEPROM} & \textbf{SSD} \\
\midrule
Suitability for ML Data & Primarily for firmware, configuration data, or small parameter sets due to limited capacity. & Suitable for large datasets, algorithms, and complex ML models due to higher capacity. \\
\midrule
ML Model Update Frequency & Less suitable for frequent updates due to wear-out constraints and limited write cycles. & Suitable for contexts requiring regular updates, attributed to better endurance and write cycles. \\
\midrule
Power Consumption & Lower, beneficial for long-duration missions where efficiency is paramount. & Higher, especially during read/write operations, which significantly impact mission power budgets. \\
\midrule
Data Access Speed & Slower read/write speeds may limit real-time ML applications. & Faster speeds enable quicker loading and execution of ML models, beneficial for real-time analytics. \\
\midrule
Reliability in Space Conditions & Highly reliable with inherent tolerance to radiation and temperature variations; requires minimal additional hardening. & Generally requires additional hardening measures due to susceptibility to SEUs, impacting data integrity for ML applications. \\
\bottomrule
\end{tabular}
\end{table}

Given the higher storage capacity of SSDs, most machine learning models for spacecraft are likely to be pre-loaded and stored using this architecture. This makes SSDs primary targets for data manipulation TTPs aimed at model poisoning or stored dataset corruption. Attackers may inject malicious data or modify existing datasets, exploiting the frequent write cycles of SSDs to alter the ML model's behavior subtly. However, tampering with the limited ML parameters or algorithms stored on EEPROM could still have targeted impacts, making specific model functions unreliable. For instance, if predetermined threshold values or decision boundaries were stored in EEPROM and referenced by a ML model, rotating these values could enable targeted impacts. EEPROMs are inherently less vulnerable to resource exhaustion attacks, owing to their static usage for data storage and inherently lower power consumption. This reduces their exposure to the high-frequency read/write operations typically exploited in such attacks. Storage exhaustion is of especial concern for SSDs in comparison to EEPROMs, given their role in storing full models. An adversarial tactic could involve intentionally inflating a convolutional neural network (CNN)'s size by adding unnecessary layers and retraining the model, thereby creating an oversized model that consumes excessive storage space and degrades system performance.

Various next-gen storage architectures for spacecraft are at their conception and will be expected to be operational post-Artemis IV in approximately 2028. For instance, Houston-based company Axiom Space has entered agreements with Kepler Communications US Inc. and Skyloom Global Corp. to build the world’s first scalable, cloud technology-enabled, commercial orbital data center (ODC) to be hosted on Axiom Station \cite{axiomspaceAxiomSpace}. One of the key features of the ODC is “Earth independence” – the ability to provide in-space cloud services without the need to connect back to terrestrial cloud infrastructure. The first tranche layer of the ODC will provide unprecedented data storage and processing capacity in a commercial, scalable, and economical way to aid satellites in LEO, MEO, and GEO through optical links via an extended mesh network \cite{axiomspaceAxiomSpace}. ODCs are likely to provide the ability for spacecraft to offload data for processing and storage, rather than solely relying on onboard storage capabilities. This capability would lead to an expanded attack surface due to the ability to poison data in the ODC which may be employed by a spacecraft to store large models. Such an attack would manifest akin to the aforementioned federated learning example whereby a poisoned model in a poorly secured node could be crosslinked into the vehicle.

\subsubsection{C\&DH Accessibility (Uplink/Write Access)}
The potential for an adversary to directly alter a machine learning (ML) model aboard a spacecraft represents a significant escalation in its risk profile. This ability is predicated on obtaining uplink and authorization access to the spacecraft's command and data handling (C\&DH) subsystem, which would allow for the deployment of software updates or the execution of specific commands. This access not only facilitates direct alterations to ML models but also enables adversaries to conduct white box attacks by thoroughly inspecting the model. Spacecraft engineers do not intend to enable C\&DH access to unauthorized individuals by design, however, vulnerabilities in ground and space segments can inadvertently grant such access. 

The advent of new ground station infrastructure paradigms, such as Ground Station as a Service (GSaaS) \cite{amazonSatelliteService}, exacerbates this threat by providing remote users with access to ground stations, thereby increasing the likelihood of ground station intrusions. Adversaries with C\&DH access are more likely to execute blunt force attacks, including direct model poisoning or complete purging of the model, rather than engaging in passive, less direct methods like black box evasion attacks. This contrast in attack methodologies is further illustrated by comparing the direct interference with a spacecraft's ML model to the aforementioned passive attack on an autonomous vehicle's computer vision system. In the latter, an adversary might subtly alter a stop sign with a nearly imperceptible sticker, causing the system to misidentify it as a speed limit sign. Such an attack manipulates the model's input data indirectly, without tampering with the model itself. In stark contrast, gaining C\&DH access to a spacecraft allows an attacker to directly compromise the ML model by injecting a malicious dataset. This approach bypasses the need to manipulate environmental inputs overtime and instead undermines the model's integrity from within, altering its training data or parameters.

\subsubsection{Model Exposure and Interaction}
Section II outlines three categories AML attacks are grouped into: white box attacks, black box attacks, and transfer attacks. These categories equally dictate the AML threats that a spacecraft will be susceptible to as they do for conventional systems. White box attacks, characterized by an adversary's complete knowledge of the machine learning model, are most likely to be perpetrated by insiders associated with the entity that owns or develops the spacecraft. Insiders may leak sensitive data regarding an ML model to unauthorized entities for financial gain or other malicious purposes. Other vectors to gaining whitebox access include the exploitation of cyber and physical vulnerabilities within the entities supply chain and IT systems to glean intelligence regarding the model. 

The extent of interaction an adversary can have with the spacecraft's ML model also indicates the type of access they can manage to gain. For instance, if an adversary can query the model and observe its outputs, they could infer sensitive information about the model's structure, its decision-making process, or the data it was trained on, effectively enabling a black box attack. This is particularly relevant if the spacecraft's ML model is accessible through an interface that allows for external queries, such as a diagnostic or maintenance port designed for remote monitoring and updates. 

Transfer attacks in the context of targeting spacecraft are likely to occur as a result of the aerospace industry's collaborative nature. ML models for spacecraft are often trained on similar datasets, especially for standardized tasks like object detection (e.g., avoiding space debris) or environmental monitoring. An adversary could exploit this by crafting input data that is known to cause misclassifications or erroneous outputs in one model, and then apply similar or identical inputs to target another spacecraft's model, anticipating comparable weaknesses. 

Historically, spacecraft benefited from an inherent defense mechanism: they were bespoke, with highly fine-tuned software making it unlikely for an attack on one system to be effective on another. However, there has been a shift towards leveraging open-source flight software frameworks, such as the NASA Core Flight System (cFS) \cite{nasaCoreFlightWebsite}, to streamline development. The adoption of common open-source frameworks like cFS increases the potential for transfer attacks, as spacecraft utilizing the same FSW framework may share vulnerabilities. For instance, \cite{ieeeWannaFlyApproach} identifies a target within cFS that could be exploited to enable a ransomware attack across any space vehicle engaging cFS, barring external security mechanisms. As NASA begins to publish trained ML models and applications for cFS, the likelihood of encountering transfer threats that exploit shared vulnerabilities across these modules will increase.
\begin{figure}
    \centering
    \includegraphics[width=1\linewidth]{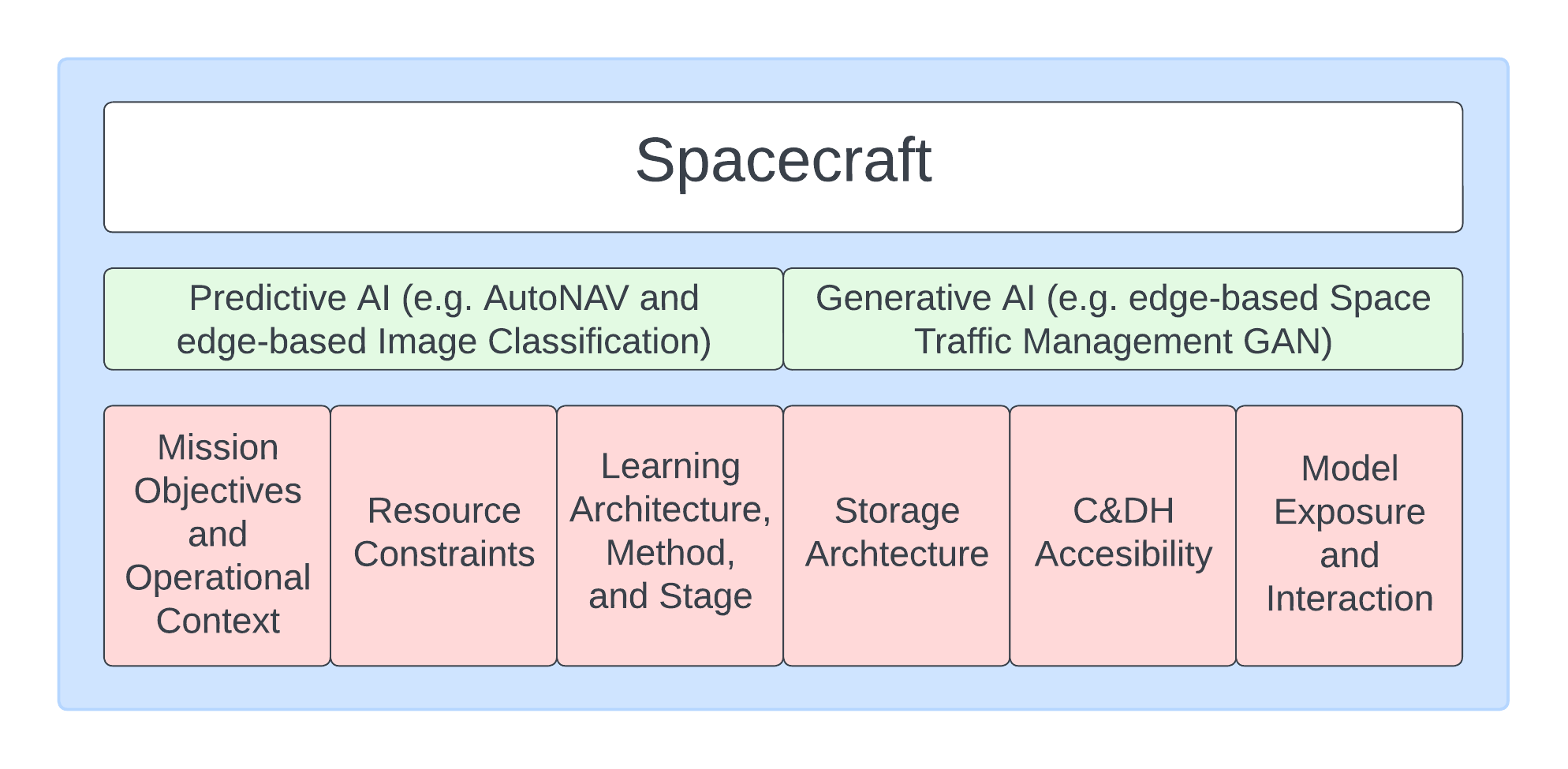}
    \caption{AML Threat Decomposition for Spacecraft}
    
\end{figure}

\section{Methods}
\subsection{NASA Core Flight System (cFS) Overview}
The NASA cFS is a generic open source flight software architecture framework utilized on flagship spacecraft, human spacecraft, and CubeSats \cite{githubGitHubNasacFS}. Developed by the NASA Goddard Space Flight Center (GSFC), cFS has been employed on a variety of missions such as the Lunar Reconnaissance Orbiter and the Global Precipitation Measurement Mission. The software framework has also been selected for NASA’s Lunar Gateway Mission in 2025 and will be essential to Gateway’s day-to-day operations \cite{nasaGoddardsCore}. cFS is designed to run on hardware as a software framework. It operates on the onboard computers (OBC) of spacecraft, satellites, and other space mission hardware, providing a runtime environment and services for custom applications developed for specific missions. The cFS framework is composed of a set of software modules that handle various tasks such as data management, communication, and command processing, making it suitable for direct implementation on space mission hardware platforms. 

\subsection{NASA On-board Artificial Intelligence Research (OnAIR) Platform Overview}
The On-board Artificial Intelligence Research (OnAIR) Platform developed by the NASA GSFC is a software framework that enables researchers to experiment with artificial intelligence algorithms in a simulated flight software environment \cite{githubGitHubNasaOnAIR}. OnAIR integrates with NASA's open source core Flight System (cFS) and supports python, which allows access to leading artificial intelligence research tools and libraries \cite{githubGitHubNasaOnAIR}.  The platform itself does not provide real satellite telemetry, algorithms for spacecraft control, or command and telemetry processing. Rather, the interface with and control of the simulated spacecraft is handled by NASA's cFS \cite{OnAirWebsiteOverview}. 

\subsection{Experimental Design}
The core objective of this research was to dissect and evaluate the potential technical strategies an adversary would employ to execute AML attacks against a spacecraft. This investigation not only aims to delineate the plausible vectors for such attacks, but also to empirically assess their potency. 

The focus for the latter was specifically laid on two types of AML attacks, poisoning and evasion, deemed most relevant for their practicality and impact on space missions. Our methodology was consequently structured around the execution of poisoning and evasion attacks in cFS, leveraging two separate datasets and corresponding models tailored for each attack scenario \footnote{https://github.com/aerospaceadversarylab/Adversarial-Machine-Learning-Threats-to-Spacecraft---Appendix}. 

\subsubsection{Attack 1 - Poisoning via Auto Navigation for GNC Algorithm Manipulation }

Autonomous GNC systems have become increasingly prevalent as engineers seek to augment legacy linear program optimizers currently employed for spacecraft dynamics. These systems rely heavily on advanced computational models to predict and adjust spacecraft trajectories in real-time, ensuring precise and safe maneuvers. To demonstrate an adversarial attack against an autonomous spacecraft navigation model, we emulated a white-box scenario whereby an adversary breached ground infrastructure to alter the integrity of the model and re-flash it onto the spacecraft's SSD. 
 
Under the public-private Tipping Point partnership between NASA and Blue Origin, two suborbital flights were performed using New Shepard to gather data on precision landing sensors \cite{newshepardarc}. New Shepard is a fully reusable, vertical takeoff, vertical landing (VTVL) space vehicle composed of two principal parts: a pressurized crew capsule and a booster rocket that Blue Origin calls a propulsion module \cite{propulsionmoduleblueorigin}. The New Shepard is controlled entirely autonomously, without ground control \cite{thespacereviewSpaceReview} or a human pilot \cite{geekwireJeffBezos}. The flights collected data from NASA-provided landing sensors, a commercial landing sensor, and host vehicle truth data for comparison \cite{newshepardarc} \cite{nasadatasetDeorbitDescent}, which constitutes this dataset. To confine the scope of this experiment, only data from the first suborbital flight was utilized. 

To operationalize the dataset, we developed a guidance, navigation, and control (GNC) program that employed stochastic gradient descent (SGD) to analyze the dataset's telemetry and sensor data to make precise predictions about the spacecraft's trajectory and control strategies required for successful deorbit, descent, and landing phases. The program first attempts to load sensor data from the dataset. This data includes various measurements like changes in velocity and angles (see dataset). After loading the data, it preprocesses it by selecting specific columns relevant for the model, and scales the features to normalize the data. The program then defines which columns of data to use as features (x) and which column to use as the target (y). In our case the target is explictly set as DATA\_DELTA\_VEL[1], implying the model will predict changes in one component of the velocity based on other velocities and angles. The program subsequently splits the preprocessed and scaled data into training and testing sets. The SGDRegressor from the scikit-learn library is then initialized and trained on the training data. The SGDRegressor uses stochastic gradient descent as the optimization algorithm to minimize the mean squared error between the predicted and actual velocity changes. This refined model would be employed by the guidance system to predict future states of the spacecraft more accurately. For example, if the model predicts that a certain combination of angle adjustments and velocity changes will lead to a desired trajectory, the guidance system can use this prediction to adjust the spacecraft's thrusters and control surfaces accordingly.
 
Malware was subsequently programmed to alter the learning rate parameter of the SGD algorithm to a value between 0.5 and 0.9, subsequently retraining the model with this compromised setting. Learning rates in this range are likely to lead to unstable training, oscillations, or divergence in the optimization process. Note that choosing to modulate the learning rate does not entail white box access. Rather, learning rate was chosen as a target due to its omnipresence in ML algorithms employed by spacecraft, such as CNNs to enable computer vision for edge-based image classification and autonomous navigation. Executing this attack in a real-world scenario would be contingent gaining the ability to re-flash the spacecraft's SSD. The mathematical representation of the SGD algorithm is defined as follows:

\[
\theta_{t+1} = \theta_{t} - \eta \cdot \nabla_{\theta} J(\theta_{t}, x_{t}, y_{t})
\]

Where $\theta$ represents the parameters of the model, $\eta$ is the learning rate, $J$ is the loss function, and $x_{t}, y_{t}$ are the input and target output at time $t$, respectively. The adversarial intervention aimed at the unauthorized rotation of $\eta$, thereby impairing the model's learning efficacy.

This attack was implemented and tested within the NASA cFS simulation environment, leveraging the NASA OnAIR Platform \cite{githubGitHubNasaOnAIR} for deploying the malware and performing the retraining phase. The results of this attack and implications are highlighted in section V. 

\subsubsection{Attack 2 - Evasion of On-Board Computer Vision through Adversarial Example Generation}
The utilization of autonomous navigation systems enabled by computer vision has become increasingly prevalent within the space industry, particularly for controlled descents onto celestial bodies. To emulate an attack against an on-board computer vision system for autonomous landing, we demonstrated an attack using genuine extracted footage from NASA's Perseverance rover landing \cite{NASA_2021}. The footage was clipped to the duration of the Terrain Relative Navigation (TRN) system's calculation of a landing solution (i.e.: safe landing spot), and split into individual frames. An image classifier trained to identify craters \cite{Mars_Crater_Study_Dataset} was employed to emulate the TRN's identification of unsafe landing spots.

To conduct the attack, a malware was written that added Poisson noise to the processed frames, prior to piping them to the model for classification. This attack would therefore not require retraining the mode, but staging a rootkit on the software bus. An overview of the attack is depicted in figure 2.

\begin{figure*}
\centering
\includegraphics[width=6.5in]{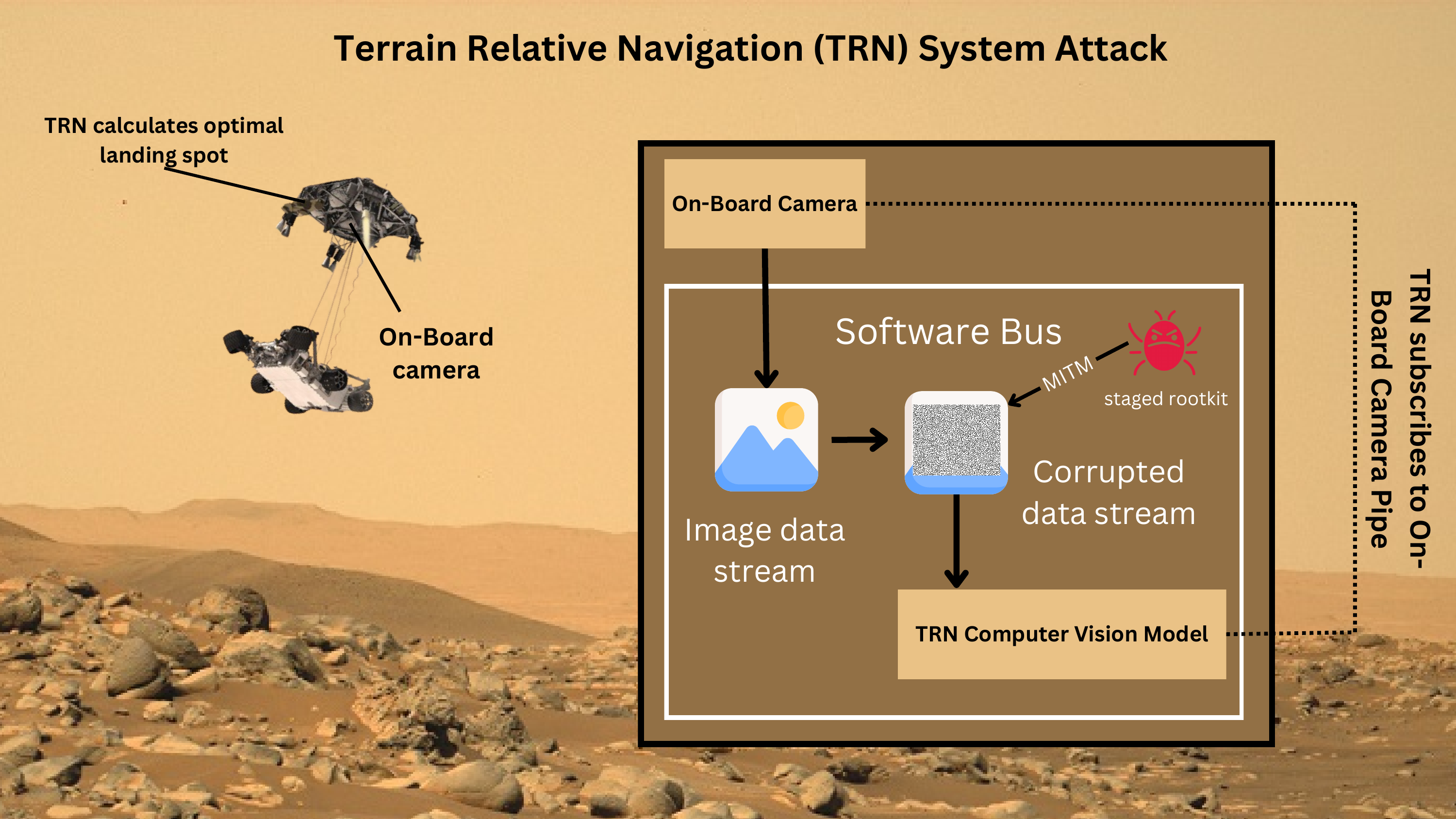}
\caption{\bf{Perseverance Rover Terrain Relative Navigation System Attack Overview}}
\label{TRN System}
\end{figure*}

When adding Poisson noise to an image, each pixel value is typically treated as a parameter $\lambda$ for a Poisson distribution, reflecting the expected number of events (e.g., photons) that would be detected by that pixel. The noisy output for each pixel is then drawn from a Poisson distribution with a mean equal to the pixel's original intensity. In mathematical terms, the noisy pixel value $y$ for an original pixel value $x$ can be modeled as:

\[
y \sim \text{Poisson}(x)
\]

Where:

\begin{itemize}
  \item $x$ represents the original pixel value (interpreted as the mean number of detected photons or similar quantifiable measures).
  \item $y$ represents the new pixel value after adding Poisson noise.
\end{itemize}

The malware processes each pixel individually whereby each pixel with intensity $x_i$, is replaced with a new value sampled from a Poisson distribution with mean $x_i$. This is modeled as:

\[
y_i = \text{random.Poisson}(x_i)
\]

Where $y_i$ is the noisy pixel value and $x_i$ is the original pixel value for each pixel $i$ in the image. By introducing noisy pixels into the image, this  caused the degradation of the crater classifier, which was subsequently unable to identify whether there was a crater at a certain location or not. This data is presented in section V. 

\section{Experiment and Discussion}

\subsection{Attack 1 Results}
The attack's efficacy was quantified by observing the mean squared error (MSE) rate---a measure of prediction accuracy---across different learning rates for both the standard (unpoisoned) and poisoned versions of the GNC algorithm.  A heightened MSE indicates a divergence between the algorithm's predictions and actual navigational states, signaling a decrease in navigational precision. Figure 2 and Table 2 encapsulate the empirical results of the nominal GNC algorithm. 

\begin{figure}
    \centering
    \includegraphics[width=1\linewidth]{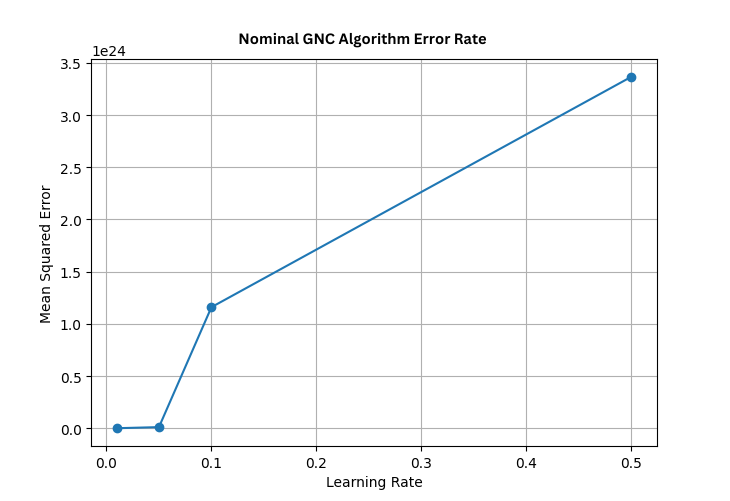}
    \caption{Baseline Performance for Nominal GNC Algorithm }
    \label{fig:baseline performance} 
\end{figure}

\begin{table}
\centering
\caption{Mean Squared Error Rate for Nominal GNC Algorithm}
\label{table:first-label1}
\begin{tabular}{|c|c|}
\hline
\textbf{Learning Rate} & \textbf{$\approx$ MSE}\\
\hline
0.001 & $0$ \\
0.050 & $0.03 \times 10^{24}$\\
0.100 & $1.28 \times 10^{24}$ \\
0.500 & $3.46 \times 10^{24}$ \\
\hline
\end{tabular}
\end{table}
In general, learning rates that exceed 0.1 cause the optimization process to overshoot and fail to converge to a good solution, as seen in Figure 2 and Table 2. However, this can vary based on a multitude of factors. This range was chosen to depict the general uptake in MSE as a result of raising the learning rate. The effect on the algorithm after executing the malware, which surreptitiously escalates the learning rate to a value between 0.5 and 0.9, can be seen in Figure 3 and Table 3. 
\begin{figure}
    \centering
    \includegraphics[width=1\linewidth]{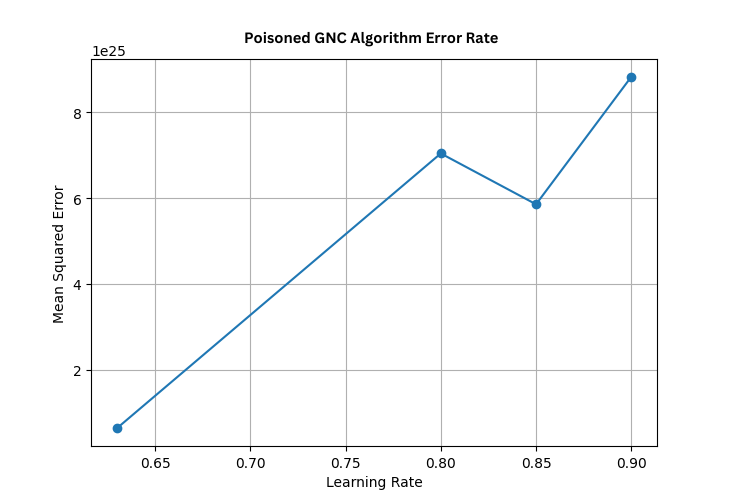}
    \caption{Performance of Poisoned GNC Algorithm}
    \label{fig:second-label} 
\end{figure}

\begin{table}[H]
\centering
\caption{Mean Squared Error Rate for Poisoned GNC Algorithm}
\label{table:second-label1}
\begin{tabular}{|c|c|}
\hline
\textbf{Learning Rate} & {$\approx$ MSE}\\
\hline
0.62 & $1.2 \times 10^{25}$ \\
0.80 & $7.3 \times 10^{25}$ \\
0.85 & $5.9 \times 10^{25}$ \\
0.90 & $8.8 \times 10^{25}$ \\
\hline
\end{tabular}
\end{table}
Post-attack, the poisoned GNC algorithm exhibited drastically elevated MSE values due to the learning rate rotation. These figures not only signify a severe decline in predictive accuracy but also translate to a total compromise in spacecraft navigational integrity. The highest MSE, reaching nearly $8.8 \times 10^{25}$, emphasizes the impact such an attack could have on mission-critical operations. Elevated MSE levels in the poisoned algorithm context directly correlate with increased risks of navigational errors, which can lead to off-course trajectories, compromised landing accuracy, or, in the worst-case scenario, mission failure. A GNC algorithm compromised by data poisoning could fail to correctly execute deorbit maneuvers, accurately navigate descent pathways, or safely manage landing operations, posing a significant risk to the spacecraft's integrity and overall mission success. 

\subsection{Attack 2 Results}
The Poisson attack against the terrain navigation system resulted in a sharp decrease in the accuracy of the model as a result of taking in perturbed and noisy images. The nominal performance of the TRN is featured in figure 4. The program was able to correctly identify unsafe landing locations with greater than 50\% assurance. After deploying the attack, however, this accuracy was decimated. 

\begin{figure}
    \centering
    \includegraphics[width=1\linewidth]{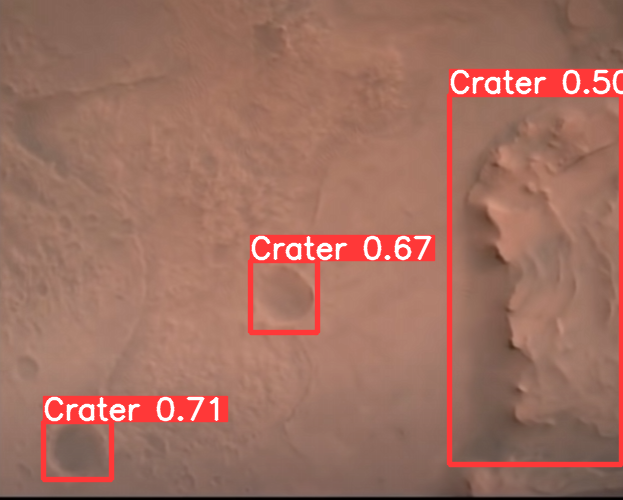}
    \caption{Performance of Nominal Terrain Navigation System} 
    \label{fig:enter-label4}
\end{figure}
Figure 5 features the performance of the TRN system after the attack. As depicted in the figure, the TRN system was unable to identify the craters due to the excessive Poisson noise introduced.  
\begin{figure}
    \centering
    \includegraphics[width=1\linewidth]{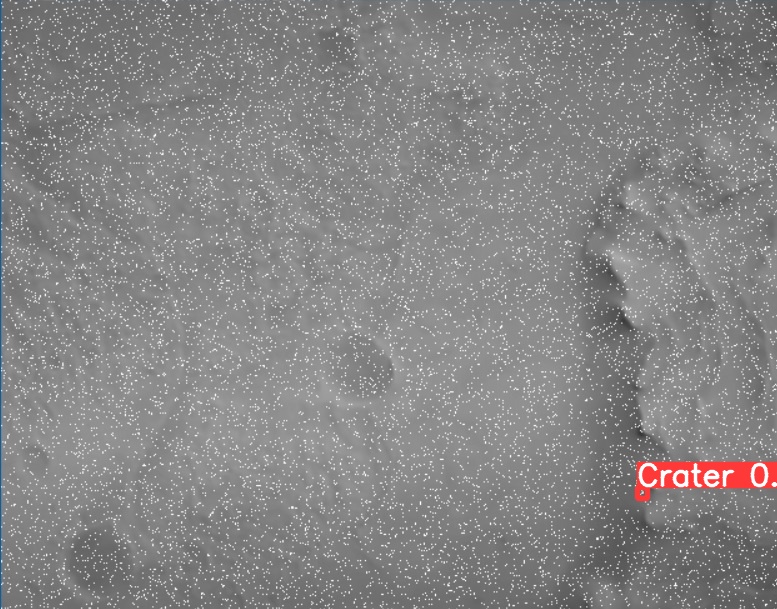}
    \caption{Performance of Attacked Terrain Navigation System}
    \label{fig:enter-label5}
\end{figure}

The introduction of these adversarial inputs into the spacecraft's operational data stream is a pivotal step to execute this attack in a real-world scenario. Such an attack would likely take the form of an advanced persistent threat rootkit that stages itself on the cFS software bus. This would enable a man-in-the-middle attack that would violate the integrity of the individual frames in the data stream between the imaging sensors and the TRN model. Upon ingestion and processing by the TRN, these altered inputs would result in erroneous classifications.

The repercussions of such misclassifications are manifold. In the domain of navigation, they could precipitate erroneous trajectory computations, misguided obstacle avoidance maneuvers, or flawed landing site selection, all of which could jeopardize the integrity of the spacecraft and the mission at large. For scientific exploratory missions, the integrity of the data collected could be compromised, potentially leading to spurious interpretations of planetary geology or the omission of significant discoveries. Lunar missions tasked with resource identification might find themselves allocating exploration assets based on fallacious data, thereby misdirecting the mission's efforts and resources. While the attack performed in this experiment employed Poisson noise for simplicity, other more covert methods could have been employed such as the fast gradient sign method (FGSM), which would perturb the inputs in a visually imperceptible manner to mission operators. 

\section{Mitigations}
Despite the nascent nature of this threat, it is imperative for spacecraft engineers to proactively integrate AML security measures as foundational design constraints for critical subsystems. By prioritizing AML security at the onset of the design and engineering processes, it is possible to circumvent the complexities associated with retroactively addressing vulnerabilities. It is also paramount to integrate AML security into vulnerability assessment frameworks for spacecraft. For each of the aforementioned parameters, specific vulnerabilities should be identified that could be exploited by AML attacks. This includes both technical vulnerabilities (e.g., in software and hardware) and operational vulnerabilities (e.g., during mission-critical events). Cross-parameter analysis should be subsequently performed  to understand how vulnerabilities in one area could exacerbate risks in another. For example, how resource constraints might limit the effectiveness of countermeasures against storage-based AML attacks. Vulnerabilities should be prioritized based on their potential impact on mission-critical systems and objectives. This helps focus mitigation efforts on areas where AML attacks could have the most severe consequences. 

Engineers are encouraged to adopt a layered security strategy that incorporates zero trust principles. Standard cybersecurity practices and mechanisms must be implemented in the ground segment such as robust non-repudiation capabilities and supply chain integrity. The spacecraft itself should have authentication measures, with multi-factor authentication capabilities preferred.  It's also advisable to avoid the employment of open-source models for spacecraft directly from the internet, as these can present unknown vulnerabilities. Keeping detailed knowledge about the operational models confidential can significantly reduce the risk of white-box attacks. Adopting standard AML security measures such as input preprocessing to detect and mitigate the effect of adversarial inputs, applying model regularization to decrease sensitivity to small perturbations, and continuously monitoring model performance for any anomalous behavior is crucial. 

It is especially paramount that spacecraft watchdogs, anomaly detection systems, and intrusion prevention systems are trained to detect atypical perturbations in model performance and can provide an early warning system against potential AML attacks. Spacecraft engineers are encouraged to consult with autonomous vehicle (automotive) companies to leverage insights from AML challenges they've faced, due to shared vulnerabilities as embedded systems. 

\section{Conclusion and Future Work}
Future directions for this work include enhancing the fidelity of experiments from NASA FSW simulations to full-scale digital twins to identify optimal mitigations. AML is projected to emerge as a potent counterspace capability as spacecraft progressively harness the advantages provided by machine learning. The evolution of ML capabilities for spacecraft is anticipated, driven by the development of new learning methods, GPUs, and architectures specifically tailored to address the unique constraints of spacecraft operations and the space environment. This evolution is expected to alter the landscape of AML threats, necessitating continuous efforts to develop effective mitigations. The taxonomy presented in this study, although not exhaustive, represents an initial effort to acclimate spacecraft engineers to the perils associated with AML.

\bibliographystyle{plain} 
\bibliography{citations}  


\end{document}